# Partial-Hessian Strategies for Fast Learning of Nonlinear Embeddings

**Max Vladymyrov** MVLADYMYROV@UCMERCED.EDU
**Miguel Á. Carreira-Perpiñán** MCARREIRA-PERPINAN@UCMERCED.EDU
Electrical Engineering and Computer Science, School of Engineering, University of California, Merced

## Abstract

Stochastic neighbor embedding (SNE) and related nonlinear manifold learning algorithms achieve high-quality low-dimensional representations of similarity data, but are notoriously slow to train. We propose a generic formulation of embedding algorithms that includes SNE and other existing algorithms, and study their relation with spectral methods and graph Laplacians. This allows us to define several partial-Hessian optimization strategies, characterize their global and local convergence, and evaluate them empirically. We achieve up to two orders of magnitude speedup over existing training methods with a strategy (which we call the *spectral direction*) that adds nearly no overhead to the gradient and yet is simple, scalable and applicable to several existing and future embedding algorithms.

We consider a well-known formulation of dimensionality reduction: we are given a matrix of $N \times N$ (dis)similarity values, corresponding to pairs of high-dimensional points $\mathbf{y}_1, \dots, \mathbf{y}_N$ (objects), which need not be explicitly given, and we want to obtain corresponding low-dimensional points $\mathbf{x}_1, \dots, \mathbf{x}_N \in \mathbb{R}^d$ (images) whose Euclidean distances optimally preserve the similarities. Methods of this type have been widely used, often for 2D visualization, in all sort of applications (notably, in psychology). They include multidimensional scaling (originating in psychometrics and statistics; Borg & Groenen, 2005) and its variants such as Sammon's mapping (Sammon, 1969), PCA defined on the Gram matrix, and several methods recently developed in machine learning: spectral methods such as Laplacian eigenmaps (Belkin & Niyogi, 2003) or locally linear embedding (Roweis & Saul, 2000), convex formulations such as maximum variance unfolding (Weinberger & Saul, 2006), and nonconvex formulations such as stochastic neighbor embedding (SNE; Hinton & Roweis, 2003) and its variations (symmetric SNE, s-SNE: Cook et al., 2007; Venna & Kaski, 2007; $t$-SNE: van der Maaten & Hinton, 2008); kernel information embedding (Memisevic, 2006); and the elastic embedding (EE; Carreira-Perpiñán, 2010). Spectral methods have become very popular because they have a unique solution that can be efficiently computed by a sparse eigensolver, and yet they are able to unfold nonlinear, convoluted manifolds. That said, their embeddings are far from perfect, particularly when the data has nonuniform density or multiple manifolds. Better results have been obtained by the nonconvex methods, whose objective functions better characterize the desired embeddings. Carreira-Perpiñán (2010) showed that several of these methods (e.g. SNE, EE) add a point-separating term to the Laplacian eigenmaps objective. This causes improved embeddings: images of nearby objects are encouraged to project nearby but, also, images of distant objects are encouraged to project far away.

However, a fundamental problem with nonconvex methods, echoed in most of the papers mentioned, has been their difficult optimization. First, they can converge to bad local optima. In practice, this can be countered by using a good initialization (e.g. from spectral methods), by simulated annealing (e.g. adding noise to the updates; Hinton & Roweis, 2003) or by homotopy methods (Memisevic, 2006; Carreira-Perpiñán, 2010). Second, numerical optimization has been found to be very slow. Most previous work has used simple algorithms, some adapted from the neural net literature, such as gradient descent with momentum and adaptive learning rate, or conjugate gradients. These optimizers are very slow with ill-conditioned problems and have limited the applicability of nonlinear embedding methods to small datasets; hours of training for a few thousand points are typical, which rules out interactive visualization and allows only a coarse model selection.

Our goal in this paper is to devise training algorithms that are not only significantly faster but also scale up to larger datasets and generalize over a family of embedding algorithms (SNE, $t$-SNE, EE and others). We do this not by simply using an off-the-shelf optimizer, but by understanding the common structure of the Hessian in these algo-rithms and their relation with the graph Laplacian of spectral methods. Thus, our first task is to provide a general formulation of nonconvex embeddings (section 1) and understand their Hessian structure, resulting in several optimization strategies (section 2). We then empirically evaluate them (section 3) and conclude by recommending a strategy that is simple, generic, scalable and typically (but not always) fastest—by up to two orders of magnitude over existing methods. Throughout we write pd (psd) to mean positive (semi)definite, and likewise nd (nsd).

## 1. A General Embeddings Formulation

Call $\mathbf{X} = (\mathbf{x}_1, \dots, \mathbf{x}_N)$ the $d \times N$ matrix of low-dimensional points, and define an objective function:

$$E(\mathbf{X}; \lambda) = E^+(\mathbf{X}) + \lambda E^-(\mathbf{X}) \qquad \lambda \geq 0 \tag{1}$$

where $E^+$ is the *attractive term*, which is often quadratic psd and minimal with coincident points, and $E^-$ is the *repulsive term*, which is often nonlinear and minimal when points separate infinitely. Optimal embeddings balance both forces. Both terms depend on $\mathbf{X}$ through Euclidean distances between points and thus are shift and rotation invariant. We obtain several important special cases:

**Normalized symmetric methods** minimize the KL divergence between a posterior probability distribution $Q$ over each point pair normalized by the sum over all point pairs (where $K$ is a kernel function):

$$q_{nm} = \frac{K(\|\mathbf{x}_n - \mathbf{x}_m\|^2)}{\sum_{n',m'=1}^N K(\|\mathbf{x}_{n'} - \mathbf{x}_{m'}\|^2)}, \quad q_{nn} = 0$$

and a distribution $P$ analogously defined on the data $\mathbf{Y}$ (thus constant wrt $\mathbf{X}$) with possibly a different kernel and width. This is equivalent to choosing $E^+(\mathbf{X}) = -\sum_{n,m=1}^N p_{nm} \log K(\|\mathbf{x}_n - \mathbf{x}_m\|^2)$, $E^-(\mathbf{X}) = \log \sum_{n,m=1}^N K(\|\mathbf{x}_n - \mathbf{x}_m\|^2)$ and $\lambda = 1$ in eq. (1). Particular cases are s-SNE (Cook et al., 2007) and $t$-SNE, with Gaussian and Student's $t$ kernels, resp. We will call $p_{nm} = w^+_{nm}$ from now on.

**Normalized nonsymmetric methods** consider instead per-point distributions $P_n$ and $Q_n$, as in the original SNE (Hinton & Roweis, 2003). Their expressions are more complicated and we focus here on the symmetric ones.

**Unnormalized models** dispense with distributions and are simpler. For a Gaussian kernel, in the elastic embedding (EE; Carreira-Perpiñán, 2010) we have $E^+(\mathbf{X}) = \sum_{n,m=1}^N w^+_{nm} \|\mathbf{x}_n - \mathbf{x}_m\|^2$ and $E^-(\mathbf{X}) = \sum_{n,m=1}^N w^-_{nm} e^{-\|\mathbf{x}_n - \mathbf{x}_m\|^2}$, where $\mathbf{W}^+$ and $\mathbf{W}^-$ are symmetric nonnegative (with $w^+_{nn} = w^-_{nn} = 0,\ n = 1, \dots, N$).

**Spectral methods** such as Laplacian eigenmaps or LLE define $E^+(\mathbf{X}) = \sum_{n,m=1}^N w^+_{nm} \|\mathbf{x}_n - \mathbf{x}_m\|^2$ and $E^-(\mathbf{X}) = 0$, with nonnegative affinities $\mathbf{W}$, but add quadratic constraints to prevent the trivial solution $\mathbf{X} = \mathbf{0}$. So $E^+$ is as in EE and SNE.

This formulation suggests previously unexplored algorithms, such as using an Epanechnikov kernel, or a $t$-EE, or using homotopy algorithms for SNE/$t$-SNE, where we follow the optimal path $\mathbf{X}(\lambda)$ from $\lambda = 0$ (where $\mathbf{X} = \mathbf{0}$) to $\lambda = 1$. It can also be extended to closely related methods for embedding (kernel information embedding; Memisevic, 2006) and metric learning (neighborhood component analysis; Goldberger et al., 2005), among others.

We express the gradient and Hessian (written as matrices of $d \times N$ and $Nd \times Nd$, resp.) in terms of Laplacians, following Carreira-Perpiñán (2010), as opposed to the forms used in the SNE papers. This brings out the relation with spectral methods and simplifies the task of finding pd terms. Given an $N \times N$ symmetric matrix of weights $\mathbf{W} = (w_{nm})$, we define its graph Laplacian matrix as $\mathbf{L} = \mathbf{D} - \mathbf{W}$ where $\mathbf{D} = \operatorname{diag}\left(\sum_{n=1}^N w_{nm}\right)$ is the degree matrix. Likewise we get $\mathbf{L}^+$ from $w^+_{nm}$, $\mathbf{L}^q$ from $w^q_{nm}$, etc. $\mathbf{L}$ is psd if $\mathbf{W}$ is nonnegative (since $\mathbf{u}^T\mathbf{L}\mathbf{u} = \frac{1}{2}\sum_{n,m=1}^N w_{nm}(u_n - u_m)^2 \geq 0$). The Laplacians below always assume summation over points, so that the dimension-dependent $Nd \times Nd$ Laplacian $\mathbf{L}^{xx}$ (from weights $w^{xx}_{in,jm}$) is really an $N \times N$ Laplacian for each $(i,j)$ point dimension. All other Laplacians are dimension-independent, of $N \times N$. Using this convention, we have for normalized symmetric models:

$$\nabla E = 4\mathbf{X}\mathbf{L} \tag{2}$$
$$\nabla^2 E = 4\mathbf{L} \otimes \mathbf{I}_d + 8\mathbf{L}^{xx} - 16\lambda \operatorname{vec}(\mathbf{X}\mathbf{L}^q) \operatorname{vec}(\mathbf{X}\mathbf{L}^q)^T$$

where $\mathbf{I}_d$ is the $d \times d$ identity matrix and we define the following scalar functions ($'$, $''$ are derivatives):

$$K = \text{kernel}, \quad K_1 = (\log K)' = K'/K, \quad K_2 = K''/K$$
$$K_{21} = (\log K)'' = (KK'' - (K')^2)/K^2 = K_2 - K_1^2$$

and weights ($K_1$ means $K_1(\|\mathbf{x}_n - \mathbf{x}_m\|^2)$, etc.)

$$w_{nm} = -K_1\,(p_{nm} - \lambda q_{nm}) \qquad w^q_{nm} = K_1\, q_{nm}$$
$$w^{xx}_{in,jm} = -\left(K_{21}\, p_{nm} - \lambda K_2\, q_{nm}\right) \times (x_{in} - x_{im})(x_{jn} - x_{jm}).$$

In particular, for s-SNE the weights are as follows:

$$w_{nm} = p_{nm} - \lambda q_{nm} \qquad w^q_{nm} = -q_{nm}$$
$$w^{xx}_{in,jm} = \lambda q_{nm}(x_{in} - x_{im})(x_{jn} - x_{jm})$$

and for $t$-SNE they are ($K$ means $1/(1 + \|\mathbf{x}_n - \mathbf{x}_m\|^2)$):

$$w_{nm} = (p_{nm} - \lambda q_{nm})K \qquad w^q_{nm} = -q_{nm}K^2$$
$$w^{xx}_{in,jm} = -(p_{nm} - 2\lambda q_{nm})(x_{in} - x_{im})(x_{jn} - x_{jm})K^2.$$

For the elastic embedding (an unnormalized model):

$$\nabla E = 4\mathbf{X}\mathbf{L} \qquad \nabla^2 E = 4\mathbf{L}\otimes\mathbf{I}_d + 8\mathbf{L}^{xx} \tag{3}$$
$$w_{nm} = w^+_{nm} - \lambda w^-_{nm} e^{-\lVert\mathbf{x}_n-\mathbf{x}_m\rVert^2}$$
$$w^{xx}_{in,jm} = \lambda w^-_{nm} e^{-\lVert\mathbf{x}_n-\mathbf{x}_m\rVert^2}(x_{in}-x_{im})(x_{jn}-x_{jm}).$$

Note the Hessian of the spectral method (i.e., for $\lambda = 0$, with constant weights $w^+_{nm}$) is constant: $\nabla^2 E = 4\mathbf{L}^+\otimes\mathbf{I}_d$.

## 2. Partial-Hessian Strategies

Our goal is to achieve search directions that are fast to compute, scale up to larger $N$, and lead to global, fast convergence. This rules out computing the entire Hessian. Carreira-Perpiñán (2010) derived pd directions for EE by using splits of the gradient such as $\nabla E = 4\mathbf{X}(\mathbf{D}^+ + (\mathbf{L}-\mathbf{D}^+)) = \mathbf{0}$ (where $\mathbf{D}^+$ is the degree matrix of $\mathbf{L}^+ = \mathbf{D}^+ - \mathbf{W}^+$), then deriving a fixed-point iterative scheme (à la Jacobi) such as $\mathbf{X} = \mathbf{X}(\mathbf{D}^+-\mathbf{L})(\mathbf{D}^+)^{-1}$ and a search direction $\mathbf{X}(\mathbf{D}^+-\mathbf{L})(\mathbf{D}^+)^{-1}-\mathbf{X}$. Here we use a more general approach that illuminates the merits of each method, by directly working with the Hessian $\nabla^2 E$. We define directions $\mathbf{p}_k\in\mathbb{R}^{Nd}$ of the form $\mathbf{B}_k\mathbf{p}_k = -\mathbf{g}_k$ where $\mathbf{g}_k$ is the gradient at iteration $k$ and $\mathbf{B}_k$ is a pd matrix (this ensures a descent direction: $\mathbf{p}_k^T\mathbf{g}_k < 0$), and use a line search on the step size $\alpha_k > 0$ satisfying the Wolfe conditions to obtain the next iterate $\mathbf{x}_{k+1} = \mathbf{x}_k + \alpha_k\mathbf{p}_k$ (Nocedal & Wright, 2006). This defines a range of methods from $\mathbf{B}_k = \mathbf{I}$ (gradient descent, very slow) to $\mathbf{B}_k = \nabla^2 E(\mathbf{X}_k)$ (Newton's direction, which would require modification to ensure descent but is too expensive anyway). We construct $\mathbf{B}_k$ as a psd part of the Hessian at $\mathbf{x}_k$ (our *partial Hessian*). Inspection of the very special structure of $\nabla^2 E$ in eqs. (2) and (3) immediately shows what parts we can use. Our driving principle is to use as much Hessian information as possible that is psd, fast to compute and leads to an efficient solution of the $\mathbf{p}_k$ linear system (e.g. sparse or constant $\mathbf{B}_k$). Note computing $E$ or $\nabla E$ is $\mathcal{O}(N^2 d)$, but solving a Hessian nonsparse linear system is $\mathcal{O}(N^3 d^3)$.

**Search directions** For normalized symmetric (and nonsymmetric) models (2), we consider functions $K$ with a nonnegative argument $t \geq 0$ and satisfying $K(t) > 0$ and $K'(t) < 0$, i.e., positive and decreasing. The term on $\mathbf{L}$ contains a psd part $-K_1 p_{nm}$ (which is constant for SNE and EE) and a nsd part $\lambda K_1 q_{nm}$; the term on $\mathbf{L}^{xx}$ is only guaranteed to contain a psd part for $i = j$ and depending on the signs of $K_2$ and $K_{21}$; and the term on $\mathbf{L}^q$ is always nsd. These psd parts can be used to construct descent directions[1]. Two important existing cases are s-SNE ($K(t) = e^{-t}$, $K_1 = -1$, $K_2 = 1$, $K_{21} = 0$) and $t$-SNE ($K(t) = \frac{1}{1+t}$, $K_1 = -K$, $K_2 = 2K^2$, $K_{21} = K^2$). For EE (an unnormalized model with $K(t) = e^{-t}$), we follow an analogous but simpler process: the hessian lacks some of the nsd parts in normalized models, e.g. the $\mathrm{vec}\,(\cdot)$ term, so it should afford better psd Hessian approximations.

**The Spectral Direction (SD)** We have found that in most cases a particular partial Hessian strikes the best compromise between deep descent and efficient computation, and yields what we call the *spectral direction* (SD). It is constructed purely from the attractive Hessian $\nabla^2 E^+(\mathbf{X}) = 4\mathbf{L}^+\otimes\mathbf{I}_d$, which as noted earlier is psd, and consists of $d$ identical diagonal blocks of $N\times N$. For EE and s-SNE this amounts to taking $\lambda = 0$ and so using the Hessian of the spectral method, thus it would achieve quadratic convergence in that case. We find it works surprisingly well for $\lambda > 0$. Effectively, we "bend" the exact gradient of the nonlinear $E$ using the curvature of the spectral $E^+$.

This basic direction is refined as follows. (1) Owing to the shift invariance of $E$, the resulting linear system is not pd but psd. To prevent numerical problems we add a small $\mu_k\mathbf{I}$ to it ($\mu_k = 10^{-10}\min(L^+_{nn})$ works well). (2) Instead of $\mathbf{B}_k\mathbf{p}_k = -\mathbf{g}_k$ (which is $\mathcal{O}(N^3 d)$) we solve two triangular systems $\mathbf{R}_k^T(\mathbf{R}_k\mathbf{p}_k) = -\mathbf{g}_k$ (which is $\mathcal{O}(N^2 d)$) where $\mathbf{R}_k$ is the upper triangular Cholesky factor of $\mathbf{B}_k$; it can be computed in place in $\mathcal{O}(\frac{1}{3}N^3)$ with standard linear algebra routines, and is sparse if $\mathbf{B}_k$ is sparse. This is crucial for scalability. For Gaussian kernels (SNE, EE) $\mathbf{L}^+$ is constant and it need only be factorized once in the first iteration. If $\mathbf{L}^+$ depends on $\mathbf{X}$, as in $t$-SNE, scalability is achieved by taking it constant (e.g. $\mathbf{L}^+$ at $\mathbf{X} = \mathbf{0}$). (3) We allow the user to sparsify $\mathbf{L}^+$ through (say) a $\kappa$-nearest-neighbor graph, which is often available as part of the data (the affinities $w_{nm}$ or probabilities $p_{nm}$). This establishes a family from $\kappa = N$ (no sparsity), which yields $\mathbf{B}_k = \mathbf{L}^+$, to $\kappa = 0$ (most sparsity), which yields $\mathbf{B}_k = \mathrm{diag}\,(\mathbf{L}^+) = \mathbf{D}^+$ (the diagonal fixed-point method of Carreira-Perpiñán, 2010).

We explored further variations in the experiments, such as updating the diagonal of $\mathbf{R}_k$ with the pd diagonal part of the full Hessian, with little improvement. Using the technique of Carreira-Perpiñán (2010) of fixed-point iteration from gradient splits, van der Maaten (2010) derives a nonsparse spectral direction for $t$-SNE, but he overlooks the fact that the resulting linear system is psd. In order to introduce spectral information during the optimization, Memisevic & Hinton (2005) use a search direction where $\mathbf{B}_k^{-1}$ (rather than $\mathbf{B}_k$) is the Laplacian. This can improve over the gradient but, as one would expect, experimentally it is not competitive with our spectral direction.

From the user point of view this yields a simple recipe that,

[1] The functions $K$ that result in the simplest Hessians would have $K_{21} = 0$ or $K_2 = 0$, which imply the Gaussian or Epanechnikov kernels, respectively. The functions $K$ that result in the Hessians having most pd parts would have $K_1 \leq 0$ (always satisfied), $K_{21} \leq 0$ and $K_2 \geq 0$; the Gaussian or Epanechnikov kernels also satisfy these conditions.

given the gradient of $E$, does not need the more complex Hessian of $E^-$. *The only user parameter is the sparsity level $\kappa$ (number of neighbors) to tune the speed of convergence; convergence itself is guaranteed for all $\kappa$ by th. 2.1.* $\kappa$ should be simply tuned to as large as computation will allow, while thresholding otherwise negligible values. The cost of computing the direction is $\mathcal{O}(N^2 d)$, the same order (less if sparse) than computing the gradient or $E$ in the line search, and we find its overhead negligible in practice. This affords directions that descend far deeper than gradient or diagonal-Hessian at the same cost per iteration.

In summary, the spectral direction works as follows. Before starting to iterate, compute the attractive Hessian $\nabla^2 E^+(\mathbf{X}) = 4\mathbf{L}^+ \otimes \mathbf{I}_d$, sparsified to $\kappa$ nearest neighbors, add the small $\mu\mathbf{I}$ to it, and cache its sparse Cholesky factor $\mathbf{R}$. At iteration $k$, given the gradient $\mathbf{g}_k$, do two backsolves $\mathbf{R}^T(\mathbf{R}\mathbf{p}_k) = -\mathbf{g}_k$ to obtain the spectral direction $\mathbf{p}_k$.

**Convergence** The following theorem guarantees global convergence (to a stationary point from any initial $\mathbf{x}_0$). It can be derived from Zoutendijk's condition and exercise 3.5 in Nocedal & Wright (2006, p. 39,63).

**Theorem 2.1.** *Consider the iteration $\mathbf{x}_{k+1} = \mathbf{x}_k + \alpha_k \mathbf{p}_k$ where $\mathbf{p}_k = -\mathbf{B}_k^{-1}\mathbf{g}_k$, $\mathbf{g}_k$ is the gradient, $\mathbf{B}_k$ is symmetric pd and $\alpha_k$ satisfies the Wolfe conditions for $k = 0, 1, 2 \dots$ If $E$ is bounded below in $\mathbb{R}^{Nd}$ and continuously differentiable in an open set $\mathcal{N}$ containing the level set of $\mathbf{x}_0$, $\nabla E$ is Lipschitz continuous in $\mathcal{N}$, and the condition number of $\mathbf{B}_k$ is bounded in $\mathcal{N}$, then $\|\nabla E(\mathbf{x}_k)\| \to 0$ as $k \to \infty$.*

In our case, we can ensure the condition number is bounded by simply adding $\mu_k \mathbf{I}$ to $\mathbf{B}_k$ with $\mu_k \geq \mu > 0$ (since $\nabla^2 E$ is bounded), which we do in practice anyway since some of our $\mathbf{B}_k$ are psd. The other conditions hold for the $E$ functions we use. From eq. (10.30) in Nocedal & Wright (2006) and with bounded condition number, it follows that

$$\|\mathbf{x}_k + \mathbf{p}_k - \mathbf{x}^*\| \lesssim r\,\|\mathbf{x}_k - \mathbf{x}^*\| + \mathcal{O}(\|\mathbf{x}_k - \mathbf{x}^*\|^2)$$

where $\mathbf{x}^*$ is a minimizer of $E$, $\mathbf{H}(\mathbf{x}^*)$ and $\mathbf{B}(\mathbf{x}^*)$ its Hessian and matrix $\mathbf{B}$, and $r = \left\|\mathbf{B}^{-1}(\mathbf{x}^*)\mathbf{H}(\mathbf{x}^*) - \mathbf{I}\right\|$. Thus the iterations have locally linear convergence with rate $r$ if we use unit step sizes (which we see in practice). The better the Hessian approximation $\mathbf{B}$ the smaller $r$ and the faster the convergence. This is quantified in the experiments.

**Other Partial-Hessians $\mathbf{B}_k$** These typically need to solve a nontrivial linear system $\mathbf{B}_k\mathbf{p}_k = -\mathbf{g}_k$. This can be accelerated in several ways: (1) by solving the system in an inexact way using linear CG initialized at the previous iteration's solution, exiting the solver after a certain tolerance $\epsilon > 0$ is achieved. (2) By updating $\mathbf{B}_k$ and its Cholesky factor every $T \geq 1$ iterations. The user has control on the exactness of the solution through $\epsilon$ or $T$. The gradient is always updated at each iteration. For the experiments in this paper we will focus on strategies with $\epsilon > 0$ and $T = 1$.

## 3. Experimental Evaluation

We have explored a number of partial Hessians as well as different strategies for efficient linear system solution, in datasets with s-SNE, $t$-SNE and EE. Here we report a representative subset of results, including what we consider the overall winner (the spectral direction). We compare the following methods: gradient descent (GD), used in SNE (Hinton & Roweis, 2003) and $t$-SNE (van der Maaten & Hinton, 2008); fixed-point diagonal iteration (FP), used in EE (Carreira-Perpiñán, 2010), much faster than GD; the diagonal of the full Hessian (DiagH); nonlinear conjugate gradients (CG) and L-BFGS (typical choices for large problems); spectral direction (SD), possibly sparsified and caching the Cholesky factor before the first iteration; and a partial Hessian $4\mathbf{L}^+ + 8\lambda\mathbf{L}^{xx}_{i*,i*}$ (which we call SD–). The latter consists of positive block-diagonal elements of $8\lambda\mathbf{L}^{xx}$ corresponding to entries associated with the same dimension ($i = j$ in $w^{xx}_{in,jm}$). This ensures a psd approximation and adds information about the Hessian of the repulsive term $E^-(\mathbf{X})$. Except for GD, FP and CG, all the other methods have not been applied to SNE-type methods that we know. Several of these methods require the user to set parameter values. For L-BFGS we tried several values for its user parameter $m$ (the number of vector pairs to store in memory) and found $m = 100$ best. For SD–, we solve the linear system with linear CG, exiting early when the relative tolerance $\epsilon$ drops below $0.1$ or we reach $50$ linear CG iterations. Generally, these parameters are hard to tune and there is little guidance on which values are the best. This is an important reason why the spectral direction, which requires no parameters to tune and performs very well, is our preferred method.

We also tried other methods that were not generally competitive and do not report them, to keep the presentation clearer. For example, adding to the SD Hessian the diagonal of the full Hessian (which depends on $\mathbf{X}$ and so varies over iterations), and solving the linear system by approximately updating the Cholesky factorization or by using CG.

Once the direction is obtained for a given method, we use a backtracking line search (Nocedal & Wright, 2006) to find a step size satisfying the first Wolfe condition (sufficient decrease). As initial step size we always try the natural step $\alpha = 1$ (recommended for quasi-Newton updates). However, we observed that some methods (in particular SD) tend to settle to accepted step sizes that are somewhat less than 1. For such cases we used an adaptive strategy: the initial backtracking step at iteration $k$ equals the accepted step from the previous iteration, $k - 1$. This is a conservative strategy because once the step decreases it cannot increase again, but it compensates in saving line searches with require expensive evaluations of the error $E$. For nonlinear CG, we use Carl Rasmussen's implementation

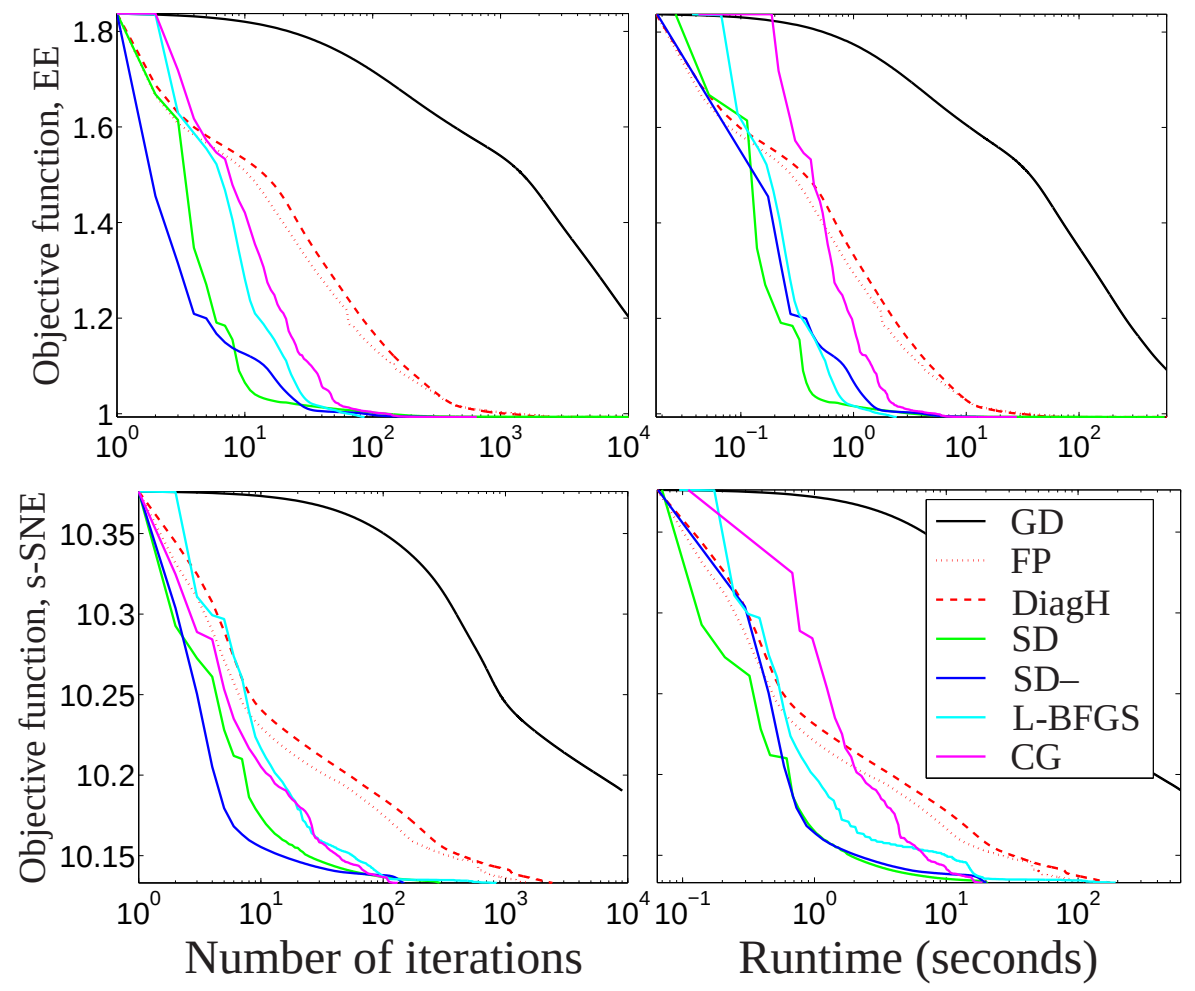

*Figure 1.* COIL-20 dataset, optimization with fixed initial and final points, for EE (top) and s-SNE (bottom). Learning curves as a function of the number of iterations (left) and runtime (right).

`minimize.m`, which uses a line search that is more sophisticated than backtracking, and allows steps longer than 1.

We evaluated these methods in a small dataset in three conditions (converging to the same minimum, converging to different minima, and homotopy training), and in a large dataset. For EE we used $\lambda = 100$.

### 3.1. Small dataset: COIL-20 image sequences

The COIL-20 dataset contains rotation sequences of objects every $5$ degrees, so each data point is a grayscale image of $128 \times 128$ pixels. We selected sequences for ten objects for a total of $N = 720$ points in $D = 16\,384$ dimensions, corresponding to ten loops (1D closed manifolds) in $\mathbb{R}^D$. In all the experiments we used SNE affinities with perplexity $k = 20$, resulting in a nonsparse $N \times N$ matrix $\mathbf{W}^+$, and reduced dimension to $d = 2$, so visual inspection could be used to assess the quality of the result. For SD we used no sparsification ($\kappa = N$).

**Convergence to the same minimum from the same initial X** We determined embeddings $\mathbf{X}_0$ and $\mathbf{X}_\infty$ such that $\mathbf{X}_\infty$ is a minimum of $E(\mathbf{X})$ and $\mathbf{X}_0$ is close enough to $\mathbf{X}_\infty$ that all methods converged to $\mathbf{X}_\infty$ when initialized from $\mathbf{X}_0$. Thus, all methods have the same initial and final destination. This allows us to reduce effects due to different local minima of the error $E$ having possibly different characteristics. Fig. 1 shows learning curves for EE and s-SNE as a function of the number of iterations and the runtime. In decreasing runtime, the methods can be roughly ordered as GD $\gg$ (FP,DiagH) $>$ (CG,SD–) $>$ (L-BFGS,SD), with GD being over an order of magnitude slower than FP, and FP about an order of magnitude slower than SD (note the log X axis). The runtime behavior and the number of iterations required agrees with the intuition that the more Hessian information the better the direction, as long as the iterations are not too expensive. Note how the method using most Hessian information, SD–, uses the fewest iterations (left panels), but these are also the slowest, which shifts its runtime curves right. For all the other methods, computing the direction costs less than computing the gradient itself. FP is very similar to DiagH. GD was the only method that did not reach the convergence value even after $10\,000$ iterations (20 minutes runtime).

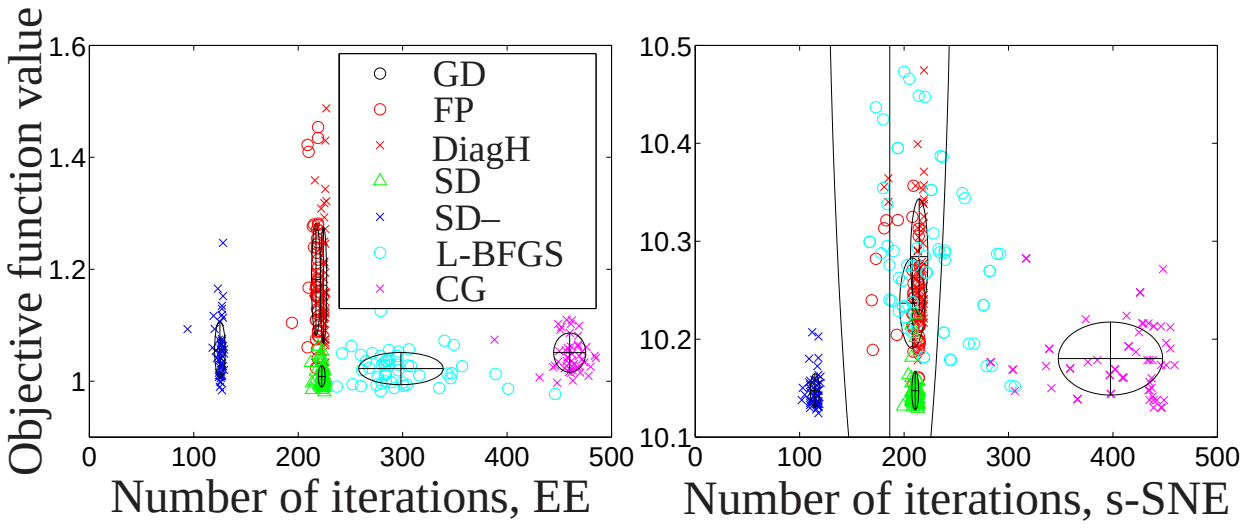

*Figure 2.* COIL-20 dataset, optimization ran for 20 s from 50 random initializations, for EE (left) and s-SNE (right).

L-BFGS is a leading method for large-scale problems. It estimates inverse Hessian information through rank-2 updates, which gives better directions than the gradient, and obtains the direction from a series of outer products rather than solving a linear system, which is fast. The main problems of L-BFGS (Nocedal & Wright, 2006, p. 180,189) are that it converges slowly on ill-conditioned problems and that, with large $Nd$, it requires an initial period of many iterations before its Hessian approximation is good. While for the small problem of fig. 1 L-BFGS is almost competitive with the SD, in the larger problem of fig. 4 it is not: 70 iterations (for EE) give a rank-$140$ approximation to a $40\text{k} \times 40\text{k}$ Hessian matrix, which fails to decrease the error.

Nonlinear CG is generally inferior to L-BFGS and this is seen in the figure too. (Our results unfairly favor CG because its `minimize.m` implementation uses a better line search than in our implementation of the other methods.)

From the beginning, the SD has an exact part of the Hessian that is pd, and obtains the direction from triangular backsolves (same cost as matrix-vector product, and dominated by the cost of computing the gradient). The only overhead is in the initial Cholesky decomposition, which is small, and progress thereafter is consistently fast.

**Convergence from random initial X to possibly different minima** We generated 50 random points $\mathbf{X}_0$ (with small values) and ran each method initialized from each $\mathbf{X}_0$, stopping after 20 seconds runtime. Fig. 2 shows the error $E$ and number of iterations for each initialization, for EE and s-SNE. They confirm the previous observations, in particular SD and L-BFGS achieve the lower errors, but SD does so more reliably (less vertical spread). GD (outside the plot) barely moved from the initial $\mathbf{X}_0$.

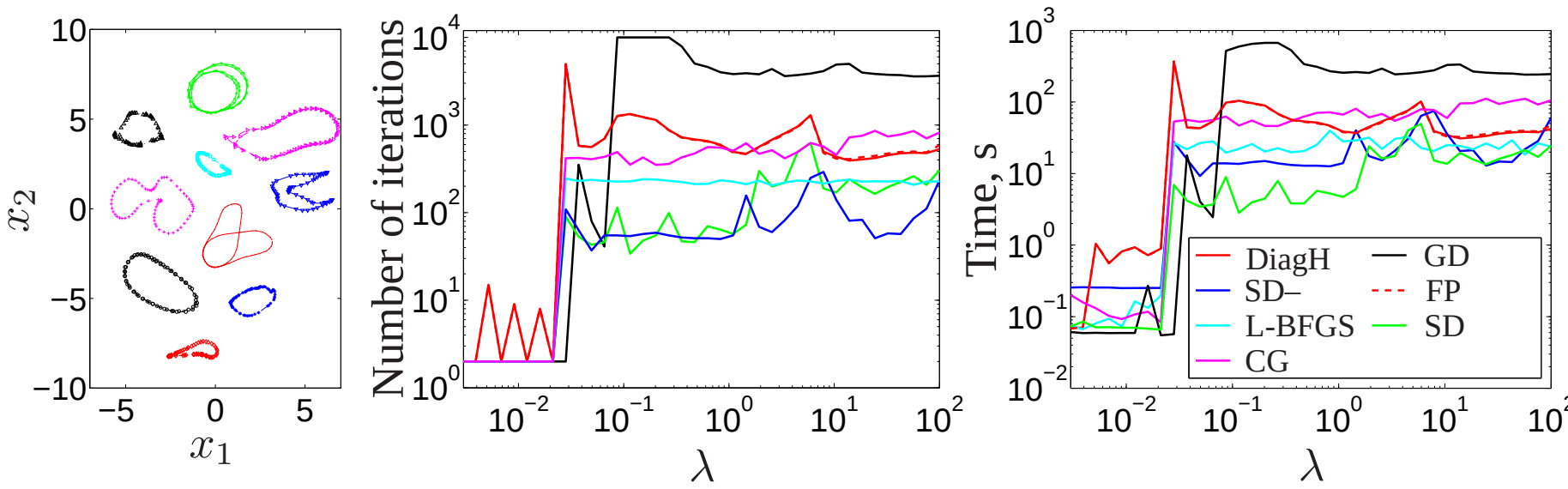

| Method | $E$ evals | Time (s) |
|---|---|---|
| GD | 143 237 | 9 291 |
| FP | 26 219 | 2 015 |
| DiagH | 26 235 | 2 016 |
| SD | 5 183 | **402** |
| SD– | **2 775** | 703 |
| L-BFGS | 6 816 | 756 |
| CG | 16 600 | 2 154 |

*Figure 3.* COIL-20. Homotopy optimization of EE. *Left*: final convergence point for nearly all methods. *Central two plots*: number of iterations and runtime required to achieve the target tolerance for each $\lambda$. *Right*: total number of error function evaluations and runtime.

**Homotopy optimization for EE** The EE error function $E(\mathbf{X};\lambda)$ can be optimized by homotopy, by starting to minimize over $\mathbf{X}$ at $\lambda \approx 0$, where $E$ is convex, and following a path of minima to a desired $\lambda$ by minimizing over $\mathbf{X}$ as $\lambda$ increases (Carreira-Perpiñán, 2010). This is slower than directly minimizing at the desired $\lambda$ from a random initial $\mathbf{X}$, but usually finds a deeper minimum. We used 50 log-spaced values of $\lambda$ from $10^{-4}$ to $10^2$ and minimized $E$ at each $\lambda$ value until the relative error decrease was less than $10^{-6}$ or we reached $10^4$ iterations. Tracking the path $\mathbf{X}(\lambda)$ so closely, we were able to have all methods converge to essentially the same embedding (shown in fig. 3), except for GD, whose embedding was still evolving (it would have required many more iterations to converge). Fig. 3 shows the runtime and the number of iterations for each $\lambda$ value.

The results again demonstrate a drastic improvement of our SD over existing methods (GD and FP from Carreira-Perpiñán, 2010), and confirm that more Hessian information results in fewer iterations and function evaluations required. Also, we observe that the SD step sizes decrease from 1 for $\lambda < 0.02$ to 0.1 for the final $\lambda$ (even though we reset to 1 the initial backtracking step every time we increase $\lambda$). Presumably, as $\lambda$ increases, so does the effect of the term $E^-(\mathbf{X})$, which the SD Hessian ignores.

### 3.2. Large dataset: MNIST handwritten digit images

While more Hessian information enables deeper decreases of the error per iteration, this comes at the price of solving a more complex linear system. To see how the different optimization methods scale up, we tested them on a dataset considerably larger than those in the literature ($N = 6\,000$ points in van der Maaten & Hinton, 2008). We used $N = 20\,000$ MNIST images of handwritten digits (each a $28\times28$ pixel grayscale image, i.e., of dimension $D = 784$). We used SNE affinities with perplexity $k = 50$ and reduced dimension to $d = 2$. All our experiments were run in a 1.87 GHz workstation, without GPUs or parallel processing. We ran several optimization methods (GD, FP, L-BFGS, SD, SD–) for 1 hour each, for both EE and $t$-SNE. For the SD we used a sparse $\mathbf{L}$ matrix with $\kappa = 7$.

As noted in section 2, for EE and s-SNE the Hessian of $E^+(\mathbf{X})$ (i.e., the matrix $\mathbf{L}^+$) is constant, so we cache its Cholesky factor before starting to iterate. For $t$-SNE, this Hessian depends on $\mathbf{X}$, and recalculating it and solving a linear system (even sparse and using linear CG) at each iteration is too costly. Thus, we fix it to the Hessian at the initial $\mathbf{X}$ and cache its Cholesky factor just as with EE. This still gives descent directions that work very well.

Fig. 4 shows the resulting learning curves for EE and $t$-SNE as a function of the number of iterations and the runtime. Some methods' deficiencies that were already detectable in the small-scale experiments become exaggerated in the larger scale. The SD– direction, while still able to produce good steps, now takes too much time per iteration (even though it is solved inexactly by CG), and is able to complete only 37 iterations for EE and 13 for $t$-SNE within the allotted time (1 hour). Note SD– does worse than SD in number of iterations even though it uses more Hessian information; this is likely due to the inexact linear system solution. In general, all methods run more iterations for EE than for s-SNE and $t$-SNE, indicating EE's simpler error function $E$ is easier to minimize. GD is omitted, because it showed no decrease of the objective function. For both EE and $t$-SNE we never observe any decrease with L-BFGS within 1 hour, although we have tried various values for $m$ ($m = 5, 50, 100$); it does decrease a little after 3 hours. This is due to the long time needed to approximate the enormous Hessian. Nonlinear CG does decrease the objective function for EE, but most of the computational resources are spent on the line search. Thus CG did least number of iterations compared to other methods. Our SD has mostly converged already in 15 minutes. SD has a reasonable setup time of 5 min. in both EE and $t$-SNE to compute the Cholesky factorization (this time can be controlled with the sparsification $\kappa$), and it is amply compensated for by the speed of the sparse backsolves in computing the direction at each iteration (which are essentially for free compared to computing the gradient). SD decreases the objective consistently and efficiently from the first iterations. FP does scale up in terms of cost per iteration, but, as in the small dataset, each step makes considerably less progress than a SD step. In summary, FP, SD– and L-BFGS are clearly not competitive with SD, which is able to scale its computational requirements and still achieve good steps.

Fig. 4 also shows the resulting embeddings for FP from (Carreira-Perpiñán, 2010) (itself much better than GD) and SD at an intermediate stage (after 20 runtime for EE and 1 hour for $t$-SNE). The difference is qualitatively obvious. The SD embedding already separates well many of the digits, in particular zeros, ones, sixes and eights. The FP embedding shows no structure whatsoever.

## 4. Discussion

Given the exceedingly long runtimes of gradient descent, we suspect some of the embeddings obtained in the literature of SNE using gradient descent could be actually far from a minimum, thus underestimating the power of SNE. The optimization methods we present, in particular the spectral direction, should improve this situation.

Experimentally, no single method is always the best. If we weigh efficiency, robustness (to user parameters) and simplicity (of implementation using existing linear algebra code and of user parameter setting), we believe that the spectral direction with cached Cholesky factor, possibly sparsified, is the preferred strategy. It achieves good steps and can be computed in less time than the gradient and objective function $E$. However, in really large problems even computing $E$ and $\nabla E$ may be too time consuming. Note that, for SNE and $t$-SNE, even if $p_{nm}$ are sparse in the attractive term, the negative term is still a full $N \times N$ matrix (though the matrix itself need not be stored for $E$ or $\nabla E$ to be computed). One solution to this is to use there a sparse graph $\mathbf{W}^-$ as in EE. However, the quality of the resulting embedding may be affected depending on the sparsity level. (Note this would not affect the construction of our spectral direction, since it does not depend on $E^-$.) Another way to accelerate the computations of sums of many Gaussians, needed in $E$ and $\nabla E$, is to use fast multipole methods (Greengard & Strain, 1991; Raykar & Duraiswami, 2006), which can reduce the time to $\mathcal{O}(N)$ if the dimension $d$ is low (which should be the case in practice).

## 5. Conclusion

We have provided a generalized formulation of embeddings resulting from the competition between attraction and repulsion that includes several important existing algorithms and suggests future ones. We have uncovered the relation with spectral methods and the role of graph Laplacians in the gradient and Hessian, and derived several partial-Hessian optimization strategies. A thorough empirical evaluation shows that among several competitive strategies one emerges as particularly simple, generic and scalable, based on the Cholesky factors of the (sparsified) attractive Laplacian. This adds a negligible overhead to the computation of the gradient and objective function but improves existing algorithms by 1–2 orders of magnitude. The quadratic cost of the gradient and objective function remains a bottleneck which future work may address. Code implementing the algorithms is available from the authors.

## Acknowledgements

Work funded in part by NSF CAREER award IIS–0754089.

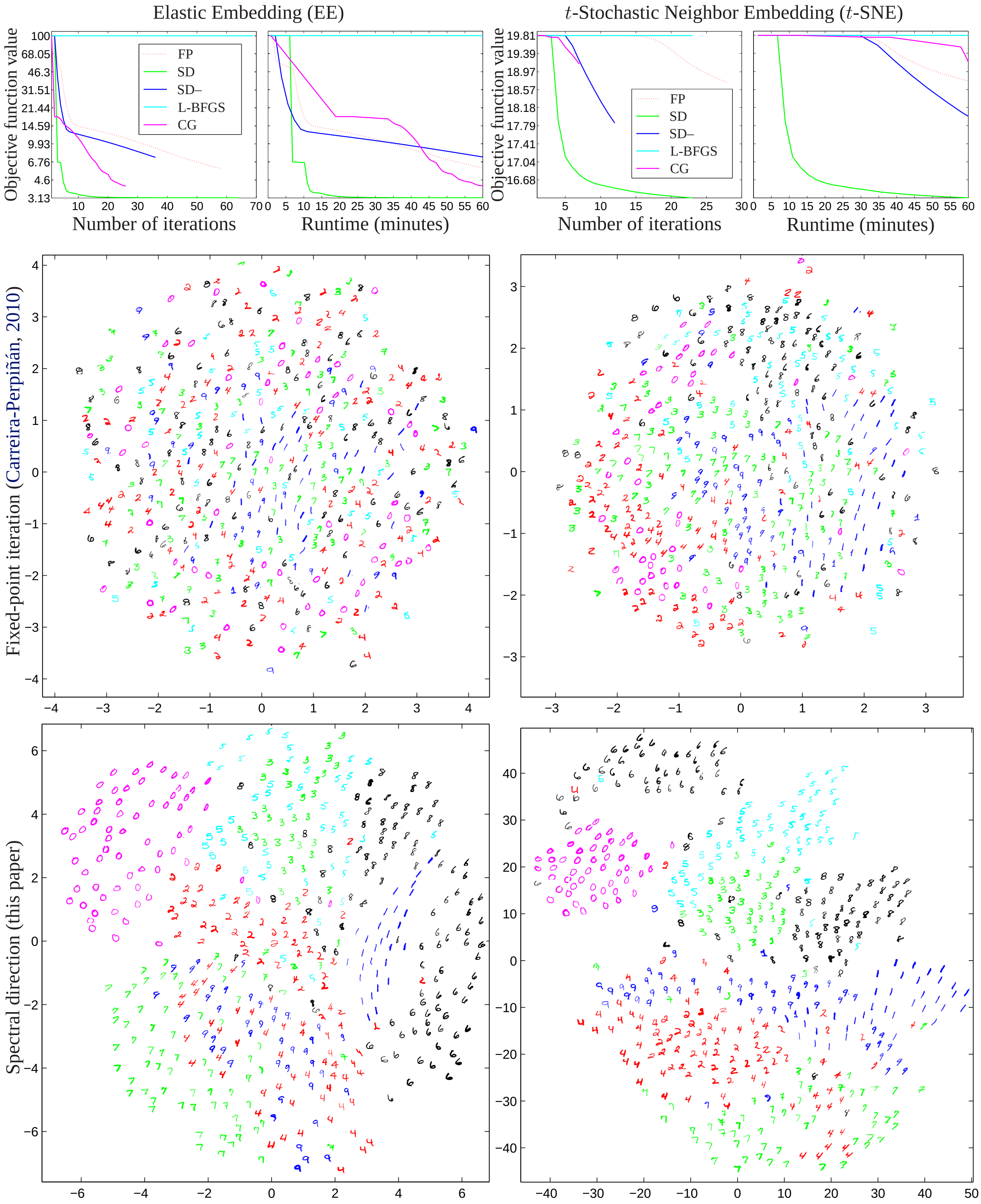

*Figure 4.* MNIST dataset, 20 000 points, for EE (left) and $t$-SNE (right). *Top*: learning curves for different methods as a function of the number of iterations (left panels) and runtime (right panels). *Bottom*: embeddings achieved after 20 minutes (EE) and 1 hour ($t$-SNE) runtime using the fixed point iteration and spectral direction. Only 500 digits shown to avoid clutter.